\documentclass[letterpaper, 10 pt, conference]{ieeeconf}
\IEEEoverridecommandlockouts                              % This command is only needed if 
\usepackage{graphics} % for pdf, bitmapped graphics files
\usepackage{epsfig} % for postscript graphics files
\usepackage{times} % assumes new font selection scheme installed
\usepackage{amsmath} % assumes amsmath package installed
\usepackage{amssymb}  % assumes amsmath package installed
\usepackage{amsmath,amsfonts} % bm
\usepackage{url}
\usepackage{hyperref}
\usepackage{comment}
\usepackage{makecell, multirow}
\usepackage{multirow}
\usepackage{booktabs}
\usepackage{amsmath}
\usepackage{array}
\usepackage{mathrsfs}
\usepackage{amssymb}
\usepackage{amsfonts}
\usepackage{amsmath}
\usepackage{cleveref}
\usepackage[flushleft]{threeparttable}
\usepackage{tablefootnote}
\usepackage{multirow}
\usepackage{hhline}
\usepackage{graphicx}
\usepackage{graphicx, subcaption}
\usepackage{hanging}
\usepackage{color}
\usepackage{tikz}
\usepackage{float,wrapfig}
\usetikzlibrary{calc,arrows,decorations.markings}
\usepackage{balance}
\usepackage{hhline}
\usepackage{algorithm}
\usepackage{mathtools}
\usepackage{svg}
\usepackage[noend]{algpseudocode}
\usepackage[sorting=none, backend=bibtex,firstinits=true,style=numeric-comp]{biblatex}

\newcommand{\printfnsymbol}[1]{%
  \textsuperscript{\@fnsymbol{#1}}%
}

\newcommand{\myparagraph}[1]{\paragraph{#1}}

\title{\LARGE \bf
Automating Robot Failure Recovery Using \\Vision-Language Models With Optimized Prompts 
}
\author{Hongyi Chen$^{1*}$, Yunchao Yao$^{1*}$, Ruixuan Liu$^{1}$, Changliu Liu$^{1}$ and Jeffrey Ichnowski$^{1}$% <-this % stops a space
\thanks{This work is supported by the Manufacturing Futures Institute at Carnegie Mellon University, funded by a grant from the Richard King Mellon Foundation. $^{*}$ denotes equal contribution.$^{1}$Hongyi Chen, Yunchao Yao, Ruixuan Liu, Changliu Liu and Jeffrey Ichnowski are with the Robotics Institute, Carnegie Mellon University {\tt\{hongyic, yunchaoy, ruixuanl, cliu6, jichnows\}@andrew.cmu.edu}
\newline
The code and prompts can be found at our GitHub repository: \url{https://github.com/hychen-naza/Robot_ErrorRecovery_VLM.git}
}
}

\begin{document}

\maketitle
\thispagestyle{empty}
\pagestyle{empty}

%%%%%%%%%%%%%%%%%%%%%%%%%%%%%%%%%%%%%%%%%%%%%%%%%%%%%%%%%%%%%%%%%%%%%%%%%%%%%%%%
\begin{abstract}
Current robot autonomy struggles to operate beyond the assumed Operational Design Domain (ODD), the specific set of conditions and environments in which the system is designed to function, while the real-world is rife with uncertainties that may lead to failures. Automating recovery remains a significant challenge. 
Traditional methods often rely on human intervention to manually address failures or require exhaustive enumeration of failure cases and the design of specific recovery policies for each scenario, both of which are labor-intensive.
Foundational Vision-Language Models (VLMs), which demonstrate remarkable common-sense generalization and reasoning capabilities, have broader, potentially unbounded ODDs.
However, limitations in spatial reasoning continue to be a common challenge for many VLMs when applied to robot control and motion-level error recovery.
In this paper, we investigate how optimizing visual and text prompts can enhance the spatial reasoning of VLMs, enabling them to function effectively as black-box controllers for both motion-level position correction and task-level recovery from unknown failures.
Specifically, the optimizations include identifying key visual elements in visual prompts, highlighting these elements in text prompts for querying, and decomposing the reasoning process for failure detection and control generation.
In experiments, prompt optimizations significantly outperform pre-trained Vision-Language-Action Models in correcting motion-level position errors and improve accuracy by 65.78\% compared to VLMs with unoptimized prompts.
Additionally, for task-level failures, optimized prompts enhanced the success rate by 5.8\%, 5.8\%, and 7.5\% in VLMs' abilities to detect failures, analyze issues, and generate recovery plans, respectively, across a wide range of unknown errors in Lego assembly.
\end{abstract}

%%%%%%%%%%%%%%%%%%%%%%%%%%%%%%%%%%%%%%%%%%%%%%%%%%%%%%%%%%%%%%%%%%%%%%%%%%%%%%%%

\section{INTRODUCTION}
% chi2023diffusion,florence2022implicit,

%Developing a general robot control policy remains a significant challenge in robotics due to the complexity of real-world environments, which inevitably leads to failures despite advances in robot learning~\cite{fang2019survey}. Collecting failure recovery demonstrations for learning-based approaches is impractical, as it would require enumerating all potential failures, many of which are unforeseen. Consequently, it is imperative to equip robots with the ability to detect and rectify their own failures.
%a capability we consider to be the cornerstone for achieving general and versatile robotic control systems.

Current robot autonomy struggles to operate beyond the assumed Operational Design Domain (ODD)—the specific set of conditions and environments for which the system is designed—due to the complexity and uncertainties of real-world environments. Outside this defined ODD, systems may not be able to guarantee safe operation, often requiring human intervention or, in some cases, ceasing to function altogether. This underscores the critical need to equip robots with the capability to detect and rectify failures~\cite{liu2023reflect, liu2024self, cornelio2024recover}.
%\cite{xiong2024aic, liu2023reflect, liu2024self, huang2022inner}.

Previous approaches, such as probability inference~\cite{diehl2022did}, scene graphs~\cite{das2021semantic}, neural-symbolic planners~\cite{cornelio2024recover}, and purely learning-based policies~\cite{inceoglu2021fino}, have been limited in their ability to address a wide range of failure scenarios and often lack generalizability. Recent efforts have focused on leveraging human feedback to correct robot failures and train policies~\cite{shi2024yell,zha2023distilling,liang2024learning,shinn2024reflexion}, but these approaches are not scalable. 

\begin{figure}[t!]
  \vspace*{-0.0in}
  \centering
    \includegraphics[width=0.47\textwidth]{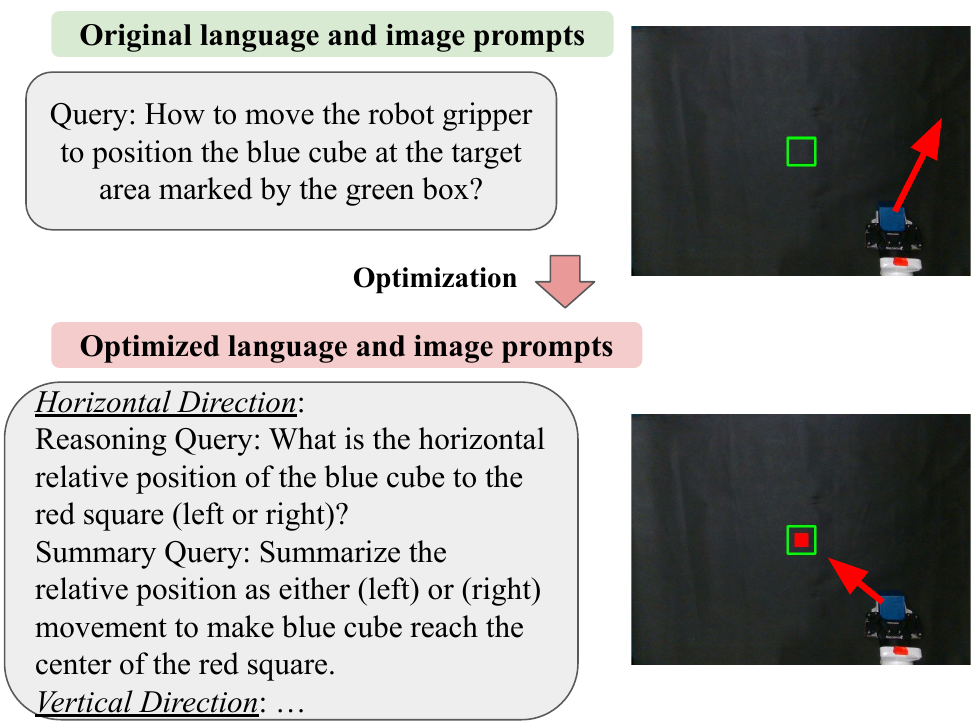}
  \vspace{-5pt}
  \caption{\small \textbf{Optimization of Language and Visual Image Input Prompts For VLMs.} The image prompts are enhanced by adding a key visual element (a red square) to highlight the goal, while the language prompts undergo optimizations in relative-position language and reasoning decomposition. The red arrow represents the movement outcome from the VLMs, demonstrating that the optimized prompts result in more accurate movement actions.}
  \vspace{-15pt}
 \label{fig:prompt}
\end{figure}

% huang2022inner
In recent years, foundational models such as Large Language Models (LLMs)~\cite{roumeliotis2023chatgpt}, Vision-Language Models (VLMs)~\cite{driess2023palm} have garnered attention from robotic researchers. These models exhibit remarkable common-sense generalization and reasoning capabilities, making them well-suited for failure explanations~\cite{liu2023reflect, shinn2024reflexion} and high-level task planning modules~\cite{ahn2022can} in robots. However, they require the conversion of multisensory observations (RGB-D, audio, robot states) into language summaries for LLM queries~\cite{liu2023reflect, shinn2024reflexion}. This process can introduce compounding errors.

% One of our goals is to enhance spatial reasoning for motion-level failure recovery without relying on model fine-tuning. 
Additionally, above foundational models struggle with 3D spatial reasoning, limiting their applicability in low-level motion planning and control~\cite{shi2024yell, chen2024spatialvlm}. 
Significant efforts and resources have been dedicated to collecting large-scale 3D spatial reasoning training data from the internet, which is used for training Vision-Language-Action (VLA) models~\cite{brohan2023rt, chen2024spatialvlm}. However, they are either closed systems~\cite{brohan2023rt} or require fine-tuning when applied to new scenarios~\cite{kim2024openvla} due to out-of-distribution challenges.
%Our work aims to simplify the translation of multisensory observations and refine both text and visual prompts to improve this process with the latest GPT-4V. 

% In this paper, we use VLMs with prompt optimizations for robot failure recovery, which get rid of the requirement of designing specific algorithm for every failure or collecting failure recovery demonstration or fine-tuning VLAs. In detail, we treat VLM as a black-box controller for failure detection and correction translating high-level RGB observations and natural language into executable recovery actions and plans, and optimize its control parameters, which are high-dimensional image prompt and language prompt, and control reasoning process. We identify the key visual elements in image prompts that the models can best analyze and highlight these elements in text prompts, which ensures the alignment of visual and textual information during model querying. We decompose the control reasoning process into multiple steps to increase the accuracy with more VLM querying cost.

In this paper, we leverage VLMs with prompt optimizations for robot failure recovery at motion-level and task-level. Our approach eliminates the need to design specific algorithms for every failure, collect failure recovery demonstrations, or fine-tune VLAs. Specifically, we treat VLMs as black-box controller for failure detection and correction, translating high-dimensional image observations and natural language into executable recovery actions and plans. We focus on optimizing VLMs control parameters, which include image prompts and language prompts, and its control reasoning process.
We identify key visual elements in image prompts that VLMs can effectively analyze, and we emphasize these elements in text prompts to ensure alignment between visual and textual information during model querying. Additionally, we decompose the control reasoning process into multiple steps, improving accuracy at the cost of increased queries (see Figure~\ref{fig:prompt}).
In summary, our contributions are as follows:
\begin{enumerate}
\item We present a study on the application of recent VLMs in detecting and recovering from unknown failures in robotics, and to the best of our knowledge, we are the first to propose a method that uses visual and text prompt optimization to enhance the spatial reasoning capabilities of VLMs.
\item In correcting motion-level position errors, VLMs with optimized prompts significantly outperform pre-trained VLAs, which struggle with tasks outside their training distribution, and achieves a 65.78\% higher accuracy compared to VLMs with original prompts.
\item Experiments in task-level Lego assembly unknown failures recovery demonstrate that the proposed prompt optimizations consistently achieve higher success rates in detecting failures, analyzing issues, and generating recovery plans compared to original prompts.
\end{enumerate}      

\section{Problem Formulation}
\begin{figure*}[t!]
  \vspace*{-0.0in}
  \centering
    \includegraphics[width=0.8\textwidth]{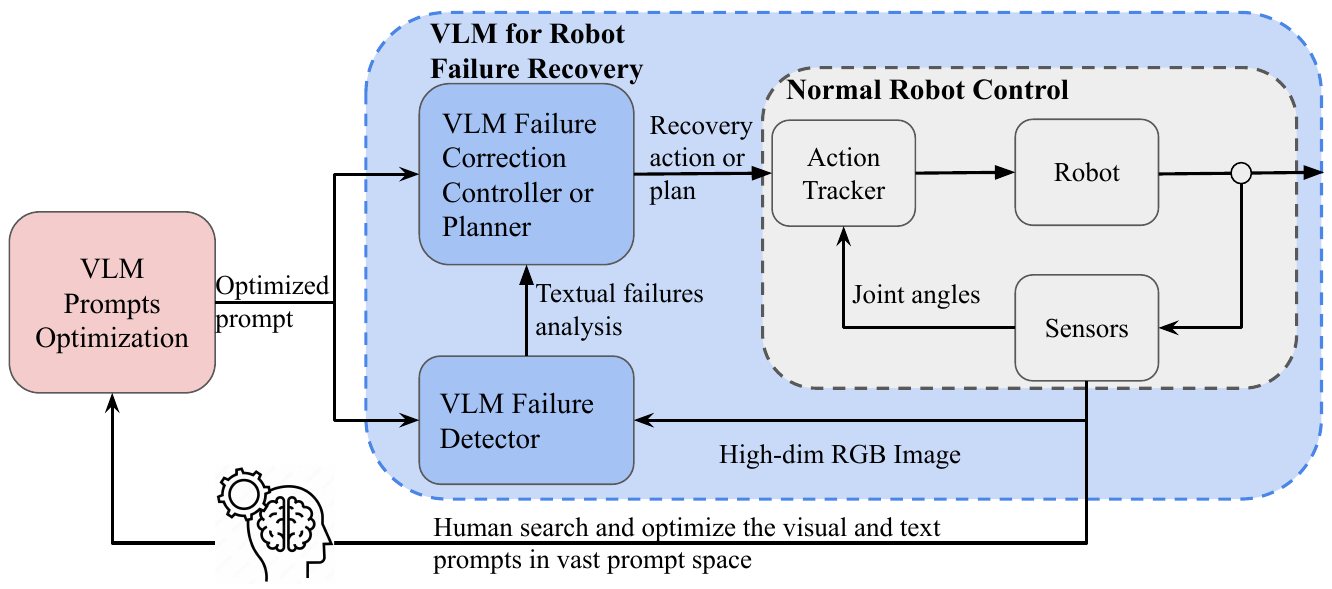}
        % \caption{Caption}
    \label{fig:pipeline}
  \vspace{-5pt}
  \caption{\small \textbf{VLMs with Optimized Prompts for Robot Failure Detection and Recovery Pipeline.} Human optimization of the VLM's input language and visual image prompts. Analyzing failure reasons to make black-box VLMs interpretable for humans and generating recovery actions or plans for robot execution.}
  \vspace{-15pt}
 \label{fig:pipeline}
\end{figure*}
Given a robot system that takes delta movement $a \in \mathbb{R}^3$ in 3D space as action input and is equipped with cameras and joint sensors for action tracking, as shown in the grey box in the inner loop of Figure~\ref{fig:pipeline}.
When controlling the robot system for a task \( T \), where an external reasoning module in the outer loop generates the control $a$, the robot may encounter two types of failures: (1) At motion level, failures may occur due to misalignment of the robot gripper with the target position, such as during grasping. (2) At task level, failures can vary significantly. For example, errors in a task like picking up a block could include failing to pick it up, grabbing the wrong color, or picking up multiple blocks.

In this paper, we aim to use a pre-trained VLM as a black-box controller \( \mathcal{M} \) to replace traditional reasoning modules for handling failures, as shown in the blue outer loop in Figure~\ref{fig:pipeline}. The VLM can be further divided into the Failure Detection module and the Failure Correction module. We propose optimizing the VLM's control parameters, which involves crafting an optimized natural language prompt \( P \) and image prompt \( I \), as well as refining the control reasoning process to maximize the accuracy of failure detection and recovery for task \( T \), as shown in the red box in Figure~\ref{fig:pipeline}. 

At motion-level, error detection and recovery requires the VLM \( \mathcal{M} \) generating delta robot motion actions $a_i \in R^3$ in 3D space or delta rotation angle $r_i \in R$ at each step depending on the task \( T \). The objective is to move the gripper as close as possible to the correct target position after the step limits $t$, utilizing VLM Failure Correction Controller with optimized prompts \( P \) and \( I \).

At task-level, we assume VLM \( \mathcal{M} \) is equipped with a basic set of robot skill actions [$\pi_1, \pi_2,..., \pi_M$], with each skill $\pi_j \in \Pi$ corresponding to a sequence of motion actions [$a_1, a_2,..., a_n$]. With the optimized prompts, the VLM must identify the reasons for the failure in VLM Failure Detector box and generate a recovery plan consisting of skills from $\Pi$ to resolve the issue in Failure Correction Planner box in Figure~\ref{fig:pipeline}.

\section{Methodology}
In this section, we detail the pipeline of using VLMs for robot failure detection and recovery, as illustrated in Fig~\ref{fig:pipeline}. The standard robot control loop, depicted within the grey box, traditionally relies on specific feature extraction algorithms or extensive policy training based on large datasets to recover failures. These approaches are often tailored to specific failure modes and struggle with high-dimensional natural language and image observations. In contrast, our approach introduces a black-box VLM correction controller that converts visual RGB images into textual descriptions of diverse failure reasons. These descriptions are then transformed into recovery actions at the motion level or recovery plans at the task level, as illustrated in the blue outer loop. Our VLM-based recovery approach extends failure recovery concepts to operate within a broader, more abstract language space and image space.

\subsection{Prompt and Reasoning Optimization for VLMs}
We propose three guidelines for optimizing the inputs—specifically language prompt \( P \), including the task description \( T \) and querying \( Q \), alongside the image prompt \( I \) provided to the VLM—and refining the query reasoning process. This process is analogous to tuning the P, I, and D parameters in a PID controller; however, instead of adjusting parameters in the numeric space, we tune the parameters of VLMs in the language and image spaces and refine the way of using these parameters to enhance spatial reasoning at the motion level and unknown failures detection at the task level.

\subsubsection{Visual Prompt Input Optimization - Key Visual Elements on Image Prompt Aligned with Language Prompt} At motion-level, it is essential to include prominent and easily noticeable visual elements for VLMs, such as distinct features on the robot gripper and the target on image prompt \( I \). If these features are not naturally present, human-added markers like colorful tape or bounding box with object detector can be used. These elements should be explicitly mentioned in language prompt \( P \) to ensure they are understandable for VLMs. As there may not be a specific definition of key visual elements at task-level, we provide an ideal reference image \( I_{ref} \) to help VLMs determine if the current skill action $\pi_i$ has been successfully implemented and identify any failures if it has not.

%For example, in the target-reaching task illustrated in Fig~\ref{fig:motion_tasks_vis}, the blue cube on the robot gripper serves as a key visual element, while the human-added red cube marker is the key element at the target. In grasping tasks in Fig~\ref{fig:motion_tasks_vis}, the blue bounding box of the detected object is the key element of the target, while the human-added red tape on the robot gripper is the key element of the gripper. The language prompt could be "How to move the red tape to the blue bounding box?" instead of "How to move the robot gripper to the water bottle?".

\subsubsection{Language Prompt Input Optimization - Relative Position-Focused Language Prompt at Motion-Level Recovery} Given the optimization of prominent and VLM-comprehensible visual elements on both the robot gripper and the target in image prompt \( I \), the language prompt \( P \) should focus on the relative positions of these key visual elements rather than the absolute positions of the original robot gripper and target. This approach is beneficial for two reasons: first, accurate reasoning about 3D world positions or delta movement can be challenging without depth, scale, and viewpoint information; second, the spatial relationship between the original robot gripper and target is more difficult to infer than the clearly indicated key elements. In this case, the continuous 3D movement action $a_i$ is transformed into a set of discrete actions $A$ = [$a_{\emph{up}}$, $a_{\emph{down}}$, $a_{\emph{left}}$, $a_{\emph{right}}$, $a_{\emph{forward}}$, $a_{\emph{backward}}$] based on relative position, each scaled by a user-defined value $c$ based on the task scenario.

%For instance, in the target-reaching task illustrated in Fig~\ref{fig:motion_tasks_vis}, the language prompt could be "What is the relative position of the blue cube to the red square and in which direction the blue cube should move to reach the red square?" instead of "How to move the blue cube to the red square?". In this case, the continuous 3D movement action $a_i$ is transformed into a set of discrete actions $A$ = [$a_{\emph{up}}$, $a_{\emph{down}}$, $a_{\emph{left}}$, $a_{\emph{right}}$, $a_{\emph{forward}}$, $a_{\emph{backward}}$], each scaled by a value $c$, which is defined by the user based on the specific task scenario.

%What is the relative position of the robot gripper to the target and in which direction the gripper should move to reach the target?".

\subsubsection{VLM Reasoning Optimization - Decomposition of VLM Query} The motivation behind decomposition is to make the failure recovery understandable to humans through textual analysis and enhance accuracy by breaking down complex recovery signal reasoning into several steps. This optimization involves two main components: \emph{(1) breaking apart VLM Failure Detector and VLM Failure Correction} and \emph{(2) breaking down spatial reasoning or failure detection inside the VLM Failure Correction module}. (1) The robot can only execute structured commands and requires strict constraints on the VLM outputs. However, because VLMs function as black-box controllers, it is difficult to gain insight into what is happening and understand the reasoning behind failures with the final structured actions. To make VLMs more interpretable, they should first fully reason about the situation using the Failure Detector module and generate a textual failure analysis. Only after this reasoning should the VLM Correction Controller or Planner summarize the actionable steps based on the reasoning results, as illustrated by the two separate blue modules in Figure~~\ref{fig:pipeline}. 

(2) At the motion level, we decompose the 3D motion-level control problems into its fundamental directional components. For example, a robot's 3D movements can be broken down into three 1D movements: up/down, left/right, and forward/backward. In this context, selecting actions from $A$ = [$a_{\emph{up}}$, $a_{\emph{down}}$, $a_{\emph{left}}$, $a_{\emph{right}}$, $a_{\emph{forward}}$, $a_{\emph{backward}}$] in a single query breaks down into its constituent 1D movements [$a_{\emph{up}}$, $a_{\emph{down}}$], [$a_{\emph{left}}$, $a_{\emph{right}}$] and [$a_{\emph{forward}}$, $a_{\emph{backward}}$] with three queries to further improve the spatial reasoning accuracy, see Sec~\ref{sec:example} for detailed examples.

Although failure types can vary at task-level, we can use human-defined correctness criteria to decompose the failure detection query and enhance analysis accuracy. For instance, instead of a single error detection query for a Lego pick-up task, such as ``The gripper is required to pick up one green Lego brick, does the gripper successfully finish the pick up action in this image?" (refer to Figure~\ref{fig:tasks_vis}), the query can be broken down into three parts: (1) ``Is the Lego block successfully picked up and attached to the robot gripper?" (2) ``Does the picked-up block have the correct color?" (3) ``Is the picked-up block a single unit or more than one block?". We have found that failures can be detected and analyzed with higher success rates by using more fine-grained queries.

\subsection{Prompt Optimization Example}
\label{sec:example}
We provide a detailed example of our step-by-step optimization process for motion-level position error recovery in the planar \emph{Target Reach} task (see Figure~\ref{fig:prompt}). First, we enhance the original visual prompt by adding a red square as a key visual element to highlight the target area, while the robot gripper is distinguished by the blue cube it grasps as its key visual element. Next, through relative-position optimization, the language prompt is refined from "How to move the blue cube to the red square?" to "What is the relative position of the blue cube to the red square, and in which direction should the blue cube move to reach the red square?" We then break down spatial reasoning into horizontal and vertical relative position reasoning, as demonstrated in the two reasoning queries shown in the optimized prompts in Figure~\ref{fig:prompt}. Finally, we separate the failure reasoning query from the recovery actions generation query using two queries. Optimized prompts for other tasks can be found at our GitHub repository.
%In \emph{Target Reach} task, as similar to other motion-level tasks, the failure reasoning focuses primarily on analyzing the relative positional relationships between the blue cube and the red square. 
%\footnote{The full prompts for all experiments can be found at our GitHub repository: \url{https://github.com/hychen-naza/Robot_ErrorRecovery_VLM.git}}.

\section{Experiments}
\subsection{Motion-level Failure Recovery Experiments}

\subsubsection{Setup} 
\begin{figure}[t!]
  \vspace*{-0.0in}
  \centering
  \includegraphics[width=0.5\textwidth]{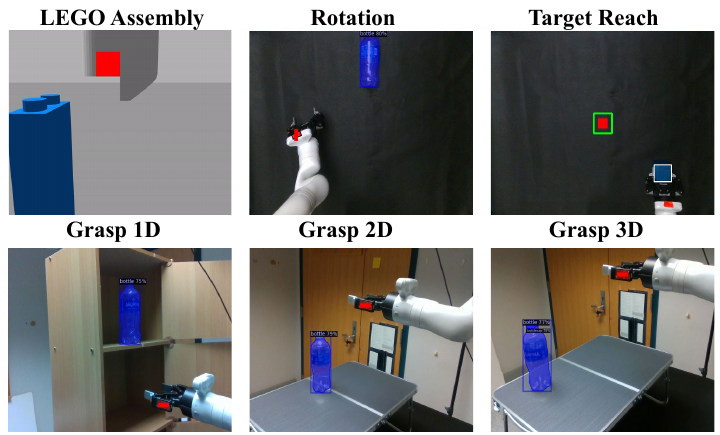}
  \vspace{-15pt}
  \caption{\small \textbf{Visualization of Four Motion-Level Tasks with Key Visual Elements.} The first row represents \emph{Lego Assembly}, \emph{Rotation} and \emph{Target Reach} tasks. The second row illustrates \emph{Object Grasp} task in 1D, 2D and 3D.}
  \vspace{-15pt}
 \label{fig:motion_tasks_vis}
 \vspace{-3pt}
\end{figure}

We conducted our evaluation on motion-level position failure recovery with GPT-4o, which takes both text and image prompts as inpu, in 7-DoF Kinova robot. We employed a zero-shot prompting approach, where GPT-4o receives no exemplars. Our evaluation focused on four robotic position failure recovery problems, where the robot is misaligned for task execution.
\begin{enumerate}
    \item \emph{Lego Assembly}: Aligning the robot gripper with the target Lego brick in simulation. The metric used is the distance to the correct position after 20 steps. 
    \item \emph{Rotation}: Rotating the gripper to align with the target object. The metric used is the angular difference from the correct angle after 10 steps. 
    \item \emph{Target Reach}: Moving the blue cube grasped by the gripper into the goal area. The metric are the coverage of the object and the green goal area and the pixel distance to the center of the goal area after 20 steps. 
    \item \emph{Object Grasp}: Moving the gripper to the correct grasping position of a water bottle in 1D, 2D, and 3D scenarios. In 1D grasping, the bottle can be placed on the top, middle, or bottom shelf, requiring the gripper to move up or down to the appropriate height for grasping. In 2D grasping, the bottle is aligned with the gripper horizontally, eliminating the need for left or right movement. The success rate of grasping after 15 steps is used as the metric.
\end{enumerate}
See Figure~\ref{fig:motion_tasks_vis} for visualization. 
We assign a step size of 0.04 m for \emph{Object Grasp} and \emph{Target Reach}, 0.002 m for \emph{Lego Assembly}, and 5° for \emph{Rotation}, reducing each value by 0.98 after every step. For comparison, we use the open-source general robot manipulation policy OpenVLA~\cite{kim2024openvla}.
%The foundational model must generate recovery controls based on available actions to move the robot to the target position. In each scenario, the robot starts at a random location, positioned at a certain distance from the intended goal. Each task is performed 10 times. 

\subsubsection{Results} 
\begin{figure*}[t!]
  \vspace*{-0.0in}
  \centering
  \includegraphics[width=0.8\textwidth]{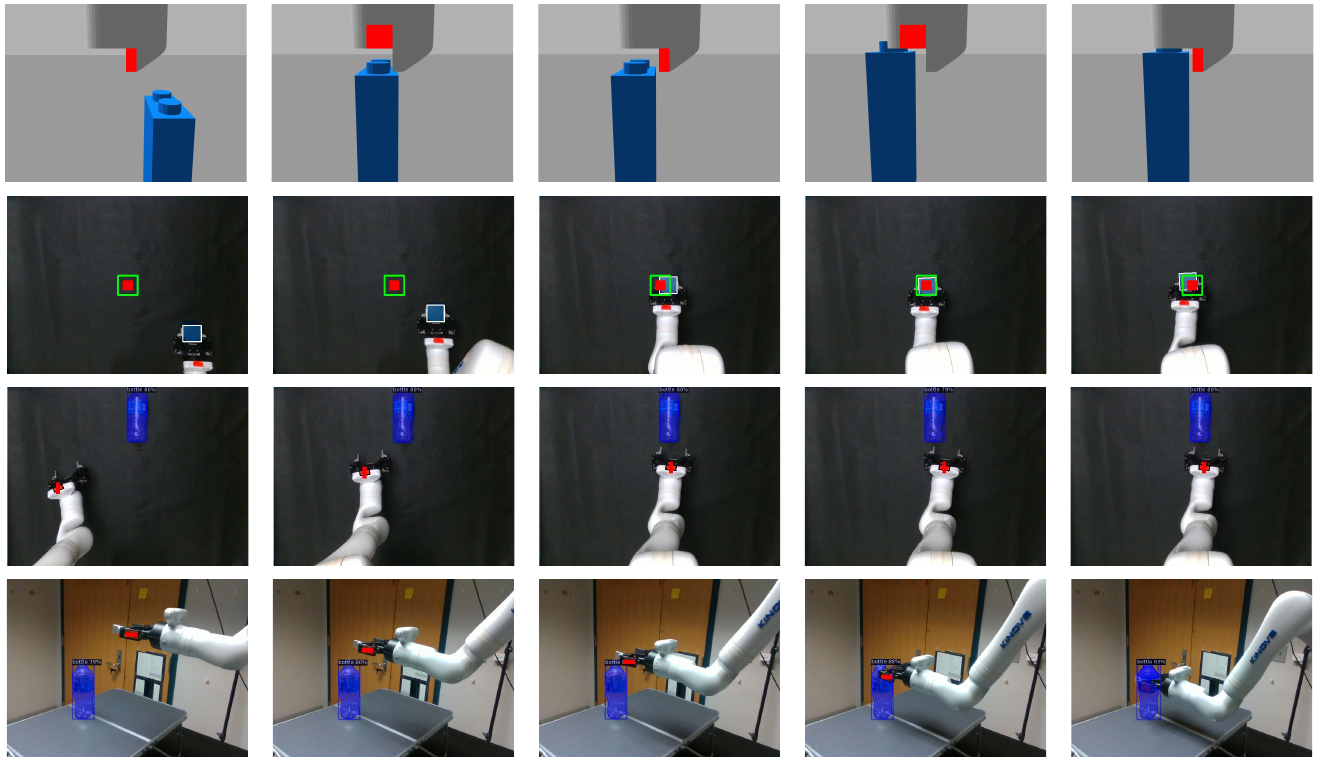}
  \vspace{-3pt}
  \caption{\small \textbf{Motion-Level Tasks Position Failure Recovery Process.} The sequence from top to bottom illustrates the tasks of \emph{Lego Assembly}, \emph{Target Reach}, \emph{Rotation}, and \emph{Grasping}, while from left to right shows the recovery process of each task.}
  \vspace{-10pt}
 \label{fig:execution}
\end{figure*}

In our experiments, we apply red tape to the robot as a visual indicator of its current position and use Detic~\cite{zhou2022detecting} to detect water bottles and generate blue bounding boxes as key visual elements for the target in the \emph{Grasp} and \emph{Rotation} tasks. Specifically in the \emph{Lego Assembly}, the red square on top determines vertical movement, while the red square on the left side determines horizontal movement, as shown in Figure~\ref{fig:execution}, to enhance performance. For \emph{Target Reach} task, we manually add a red square as the key visual element. 

After the relative position optimization, the available actions are as follows: [$a_{\emph{up}}$, $a_{\emph{down}}$, $a_{\emph{left}}$, $a_{\emph{right}}$] in the \emph{Lego Assembly}; [$a_{\emph{up}}$, $a_{\emph{down}}$] for 1D \emph{Grasp}; [$a_{\emph{up}}$, $a_{\emph{down}}$, $a_{\emph{forward}}$, $a_{\emph{backward}}$] for 2D \emph{Grasp}; and all six directions for 3D \emph{Grasp}. For the \emph{Rotation} task, the available actions are [$a_{\emph{rotateLeft}}$, $a_{\emph{rotateRight}}$], while for the Target Reach task, they are [$a_{\emph{up}}$, $a_{\emph{down}}$, $a_{\emph{forward}}$, $a_{\emph{backward}}$]. Main results are in the Table~\ref{tab:main_res}.

\emph{(1) Compared to OpenVLA, our approach is simpler to use, more effective at handling out-of-distribution problems, and achieves higher recovery accuracy.} While OpenVLA performs well in grasping tasks, its performance declines in other areas, likely because grasping tasks closely resemble OpenVLA's training settings. This suggests that current VLAs still lack the strong generalization and robustness seen in LLMs, as confirmed by its authors, who note that fine-tuning is required for new tasks and scenarios. In contrast, our approach achieves significantly higher recovery success rates and accuracy across all tasks, without the need for any fine-tuning or additional model setup, demonstrating its strong capability for broader ODDs. Refer to Figure~\ref{fig:execution} for recovery executions, we observe that GPT-4o occasionally struggles with spatial reasoning during execution, even with our optimizations, and has difficulty handling the stop condition. Thus, the final step's metric may not always reflect the optimal point along the trajectory. For example, in Figure~\ref{fig:execution} \emph{Target Reach} task, the blue cube is closer to the center at the second-to-last step than in the final step. Besides, our approach does not yet account for collisions during movement, which could knock over the water bottle and result in failure during 3D grasping.

\emph{(2) We study the contribution of each optimization techniques in \emph{Target Reach} task in Table~\ref{tab:abl_res}.} First, we transform the action representation from continuous 3D world delta movements into a set of discrete actions using relative-position optimization. This increases coverage to $4.99\%$ and reduces the pixel distance by $7$, as the blue cube occasionally moved into the target area, leaving the pixel distance relatively large. When combined with decomposition, coverage surges to $31.42\%$ and the distance drops to $34$, demonstrating the effectiveness of providing VLMs with more resources for action reasoning. However, the blue cube only partially covers the target area, indicating that accuracy remains suboptimal. Finally, by incorporating prominent key visual elements in both the image and text prompts, coverage increases to $65.78\%$, with the blue cube positioned very close to the center of the target, where slight movement corrections could further enhance coverage greatly.

%\vspace{-7pt}
\subsection{Task-level Failure Recovery Experiments}
\begin{figure*}[t!]
  \vspace*{-0.0in}
  \centering
  \includegraphics[width=0.75\textwidth]{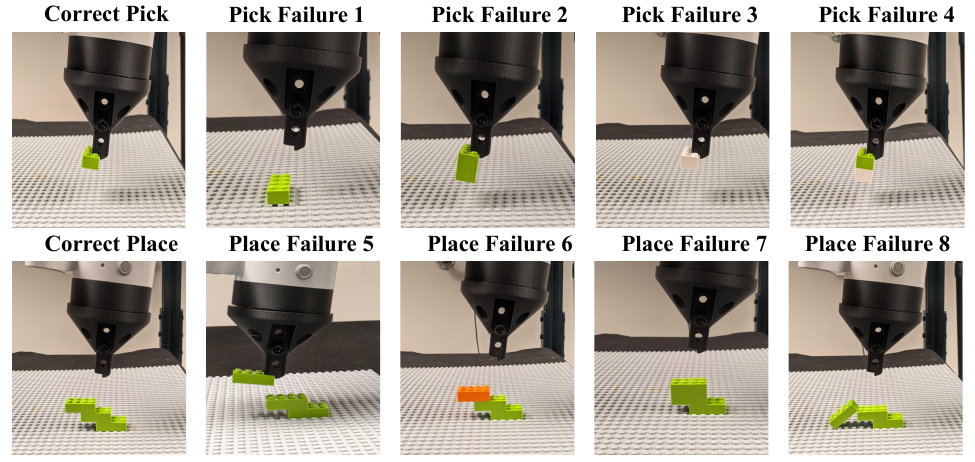}
  \vspace{-5pt}
  \caption{\small \textbf{Visualization of Eight Task-Level Failures in Lego Pick and Place.} The first row depicts four failures during the Lego pick phase, while the second row illustrates four failures during the Lego place phase.}
  \vspace{-15pt}
 \label{fig:tasks_vis}
\end{figure*}

\myparagraph{Setup} We evaluated our approach on the Lego assembly process, which can be broadly categorized into two tasks: picking and placing blocks. The evaluation utilized GPT-4o in a zero-shot prompting scenario, where a correct reference image was provided to help GPT-4o reason out the causes of failures for both original and optimized prompts. To assess our approach's ability to detect and recover from failures, we created four potential failure cases for Pick and Place respectively. \textbf{Pick}: (1) \textit{Fail to pick up block}, (2) \textit{Pick multiple blocks}, (3) \textit{Pick a block of the wrong color}, (4) \textit{Pick multiple blocks including the wrong color block}. \textbf{Place}: (5) \textit{Fail to place block}, (6) \textit{Place a block of the wrong color}, (7) \textit{Place the block in the wrong position}, (8) \textit{Collapse of the structure caused by excessive force applied during placement}. Refer to the visualizations in Figure~\ref{fig:tasks_vis} for correct and failed images. For each failure scenario, we assessed the success rate across three criteria: (1) \emph{Failure Detection}: determining whether GPT-4o identifies the failure in the image (Yes/No); (2) \emph{Failure Analysis}: assessing whether GPT-4o accurately identifies the reason for the failure as listed above; and (3) \emph{Recovery Planning}: evaluating whether GPT-4o can generate a viable recovery plan using the robot's skill actions [Pick, Place, Sweep away block, Move to pickup location, Move to place location, Move to discard location]. A correct plan should reconstruct the Lego model as shown in the reference image, assuming all skill actions are executed successfully. Additionally, certain pre-conditions must be met for specific skills, such as for Pick, where the gripper must be empty and at the pickup location, and for Place, where the gripper must have a Lego block attached and be at the place location. The Sweep action is specifically designed to clear blocks in the case of Failure 8.

\begin{table}[h]
\caption{\small \textbf{Quantitative Performance of GPT-4o with All Optimizations and Baseline Model in Motion-Level Failure Recovery.} The averaged performance across 10 random initial positions for both models. "Dist (m)" refers to the 3D spatial distance to the correct alignment position, "Angle Dist" denotes the difference between the current and correct angles, and "Coverage" means the coverage of the blue cube and the green goal area.}
\vspace{-5pt}
\begin{center}
\small
\scalebox{0.9}{
\begin{tabular}{@{}ccccccc@{}}
\toprule
\bf Model &\multicolumn{1}{c}{{\bf Lego}}& \bf Rotation & \multicolumn{1}{c}{\bf Target reach} & \multicolumn{3}{c}{\bf Object Grasping} \\
&\bf Dist & \bf Angle Dist& \bf Coverage & \bf 1D & \bf 2D & \bf 3D \\
\midrule
OpenVLA&0.016&24.3&0\%&4/10&6/10&0/10\\
Our&0.005 & 7.4 & 65.78\%&10/10& 8/10&4/10\\
\bottomrule
\end{tabular}
}
\end{center}
\label{tab:main_res}
\vspace{-10pt}
\end{table}

\begin{table}[h]
\caption{\small \textbf{Quantitative Analysis of Contribution of Each Optimization.} Performance comparison of GPT-4o using the original prompts, partially optimized prompts, and fully optimized prompts. "Pixel Dist" indicates the pixel distance between the center of the blue cube and the center of the red square.}
\vspace{-5pt}
\begin{center}
\small
\begin{tabular}{@{}ccc@{}}
\toprule
\multirow{2}*{\makecell[c]{\bf Model}} & \multicolumn{2}{c}{\bf Target reach} \\
& Coverage & Pixel dist\\
\midrule
original prompt \( P \) & 0\% & 215\\
\( P \) + relative & 4.99\% & 202 \\
%\( P \) + relative and action query decomp ($\emph{b}$+$\emph{c}_{\emph{1}}$)& 0\%&224\\
\( P \) + relative and decomposition & 31.42\% & 34\\
\( P \) + full optimizations & 65.78\% & 19\\
\bottomrule
\end{tabular}
\end{center}
\label{tab:abl_res}
\vspace{-10pt}
\end{table}
\myparagraph{Results} The main results of optimized prompt are presented in Table~\ref{tab:main_res_high}. For comparison, we also include the results for the original prompt, with scores 109 out of 120 for failure detection, 103/120 for failure analysis, and 79/120 for planning. Due to GPT-4o's already strong common-sense reasoning from image and language inputs, our optimization resulted in 6.4\% success rate improvement across the three metrics, which is less significant compared to the gains observed in motion-level tasks. Failure analysis is challenging in tasks that are visually subtle, such as \emph{Pick Failure 2}, where two blocks of the same color are involved; \emph{Pick Failure 4}, which requires detecting both the wrong color and the incorrect number of blocks; and \emph{Place Failure 7}, where the shape closely resembles the reference image. The success rate of recovery planning is lower due to the inclusion of irrelevant skill actions, which result in a Lego structure that differs from the reference image, or the omission of necessary pre-skill actions required to meet preconditions. For example, in \emph{Pick Failure 1}, the recovery plan sometimes includes a place action immediately after picking up, which exceeds the scope of the pick task, while in \emph{Place Failure 7}, the plan instructs the robot to place the block at step one, even though the gripper is empty. 

%More structured methods, such as temporal logic planning~\cite{wang2023conformal}, could potentially address these issues.

\begin{table}[h]
\vspace{-5pt}
\caption{\small \textbf{Quantitative Performance of GPT-4o with Optimized Prompt in Task-level Failure Recovery.} The success rate is evaluated across 3 unique sets, with one set illustrated in Figure~\ref{fig:tasks_vis}. Each set is tested using 5 random seeds. D means failure detection success rate; A means failure analysis success rate; P means recovery planning success rate.}
\vspace{-5pt}
\begin{center}
\small
\scalebox{0.95}{
\begin{tabular}{@{}cccc@{}}
\toprule
\bf Failure& \bf D & \bf A & \bf P \\
\midrule
% it is detecting correctly but has error in summary.
% need to add false positive matrix
%Pick Success &5/7 & 5/7 & -\\
Fail to pickup block&15/15 & 15/15 & 8/15\\
Pick multiple blocks& 13/15 & 13/15 & 7/15\\
Pick a block of the wrong color&15/15 & 15/15 & 13/15\\
{\parbox{3.cm}{Pick multiple blocks with wrong color ones }}&15/15 & 10/15 & 10/15\\
\midrule
 Fail to place block & 15/15 & 15/15 & 12/15\\
Place a block of the wrong color& 15/15  & 15/15  & 13/15\\
Place the block in the wrong position& 13/15  & 12/15  & 10/15\\
Collapse of the placed structure& 15/15 & 15/15  & 15/15\\
\midrule
\bf Summary & 116/120 & 110/120 & 88/120\\
\bottomrule
\end{tabular}
}
\end{center}
\label{tab:main_res_high}
\vspace{-15pt}
\end{table}

\section{CONCLUSIONS}
This work introduces and evaluates the use of VLMs as failure detectors and correctors for robot motion-level control and task-level planning. The proposed prompt optimization techniques have demonstrated significant improvements across a variety of tasks at both levels.

% \bibliographystyle{IEEETran}
% \bibliography{root.bib}
\printbibliography
\end{document}